\documentclass{bmvc2k}
\usepackage{times}
\usepackage{epsfig}
\usepackage{graphicx}
\usepackage{amsmath}
\usepackage{amssymb}
\usepackage{caption}
\usepackage{multirow}
\usepackage{arydshln}
\usepackage{color}
\usepackage[table,xcdraw]{xcolor}
\usepackage[ruled,vlined]{algorithm2e}

\def\thefootnote{*}\footnotetext{These authors contributed equally to this work.}

\graphicspath{{./images/}}

\title{CAMS: Color-Aware Multi-Style Transfer}

\addauthor{Mahmoud Afifi}{mafifi@eecs.yorku.ca}{1}

\addauthor{Abdullah Abuolaim\thefootnote{}}{abuolaim@eecs.yorku.ca}{1}

\addauthor{Mostafa Hussien\thefootnote{}}{mostafa.hussien@mail.mcgill.ca}{2}

\addauthor{Marcus A. Brubaker}{mab@eecs.yorku.ca}{1}

\addauthor{Michael S. Brown}{mbrown@eecs.yorku.ca}{1}

\addinstitution{
York University\\
Toronto, Ontario\\
Canada
}

\addinstitution{
\'Ecole de technologie sup\'erieure (\'ETS)\\
Montreal, Quebec\\
Canada
}

\runninghead{Afifi \etal}{CAMS: Color-Aware Multi-Style Transfer}

\def\eg{\emph{e.g}\bmvaOneDot, }

\def\etal{\emph{et al}\bmvaOneDot, }

\begin{document}

\maketitle

\begin{figure}[t]
\includegraphics[width=\textwidth]{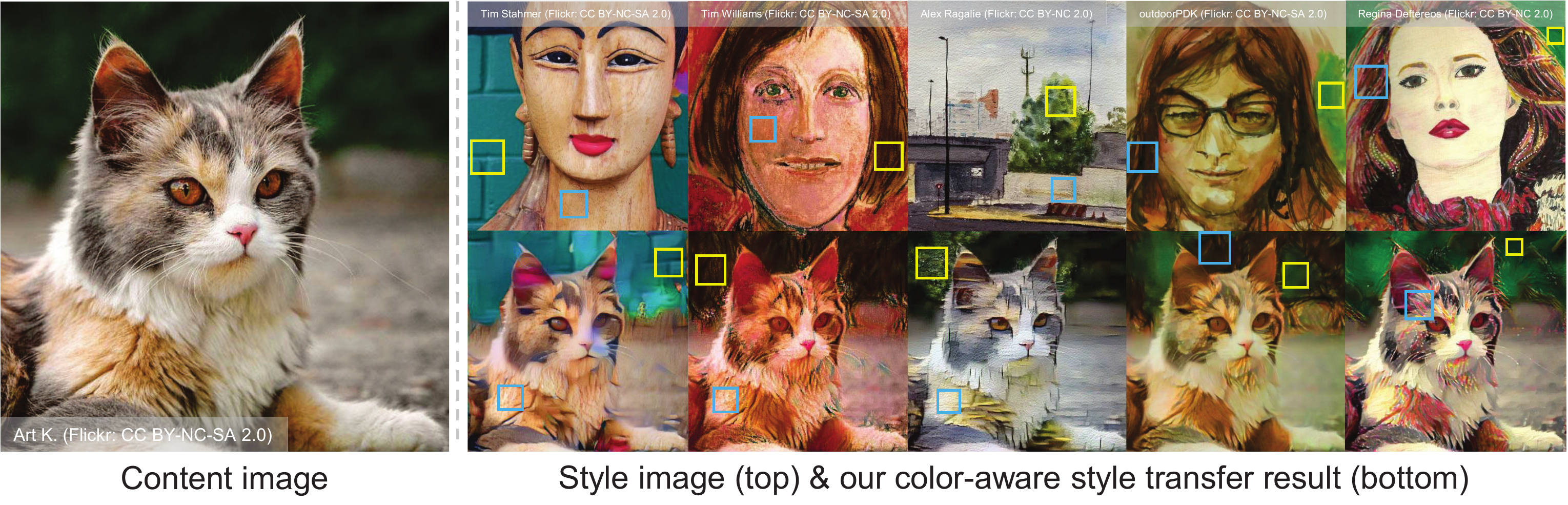}
\vspace{0.5mm}
\caption{We propose a color-aware multi-style transfer method that generates aesthetically pleasing results while preserving the correlation between the style and colors in the reference image (i.e., style image) and the generated one. Each shown style image (top) has different styles. We highlight two example styles in each style image in yellow and blue. We also highlight the corresponding transferred styles in our results (bottom).}
\label{fig:teaser}
\end{figure}

\begin{abstract}
Image style transfer aims to manipulate the appearance of a source image, or ``content'' image, to share similar texture and colors of a target ``style'' image. Ideally, the style transfer manipulation should also preserve the semantic content of the source image. A commonly used approach to assist in transferring styles is based on Gram matrix optimization. One problem of Gram matrix-based optimization is that it does not consider the correlation between colors and their styles. Specifically, certain textures or structures should be associated with specific colors.  This is particularly challenging when the target style image exhibits multiple style types.  In this work, we propose a color-aware multi-style transfer method that generates aesthetically pleasing results while preserving the style-color correlation between style and generated images. We achieve this desired outcome by introducing a simple but efficient modification to classic Gram matrix-based style transfer optimization. A nice feature of our method is that it enables the users to manually select the color associations between the target style and content image for more transfer flexibility. We validated our method with several qualitative comparisons, including a user study conducted with 30 participants. In comparison with prior work, our method is simple, easy to implement, and achieves visually appealing results when targeting images that have multiple styles. Source code is available at \href{https://github.com/mahmoudnafifi/color-aware-style-transfer}{https://github.com/mahmoudnafifi/color-aware-style-transfer}.
\end{abstract}


\section{Introduction} \label{sec:intro}
Style transfer aims to manipulate colors and textures of a given source image to share a target image's ``look and feel''. The source and target images are often so-called ``content'' and ``style'' images, respectively, where style transfer methods aim to generate an image that has the content of the source image and the style from the target image. 

One of the first deep learning methods to produce images with artistic style was Deep Dream \cite{deepdream}. Deep Dream works by reversing a convolutional neural network (CNN), trained for image classification via image-based optimization. The process begins with a noise image, which is iteratively updated through an optimization process to make the CNN predicts a certain output class. Inspired by Deep Dream, Gatys \etal~\cite{gatys2015neural} proposed a neural style transfer (NST) method to minimize statistical differences of deep features, extracted from intermediate layers of a pre-trained CNN (\eg VGG net~\cite{simonyan2014very}), of content and style images. After the impressive results achieved by the NST work in \cite{gatys2015neural}, many methods have been proposed to perform style transfer leveraging the power of CNNs (\eg \cite{gatys2016preserving, gupta2017characterizing, li2017universal, luan2017deep,snelgrove2017high, ruder2018artistic, shen2018neural, Tesfaldet2018, park2019arbitrary, yang2019controllable, yoo2019photorealistic, svoboda2020two, wang2020collaborative, wang2020diversified, yim2020filter, liu2021learning, afifi2021histogan, kotovenko2021rethinking}).

The work presented in this paper extends the idea of image-optimization NST to achieve multi-style transfer through a color-aware optimization loss (see Figure \ref{fig:teaser}). We begin with a brief review of image-optimization NST in Section \ref{sec:background}, then we will elaborate our method in Section \ref{sec:method}. 

\section{Background} \label{sec:background}
NST-based methods use similarity measures between CNN latent features at different layers to transfer the style statistics from the style image to the content image. In particular, the methods of ~\cite{gatys2015neural, berger2016incorporating} utilize the feature space provided by the 16 convolutional and 5 pooling layers of the 19-layer VGG network~\cite{simonyan2014very}. The max-pooling layers of the original VGG were replaced by average pooling as it has been found to be useful for the NST task. For a pre-trained VGG network with fixed weights and a given content image, the goal of NST is to optimize a generated image so that the difference of feature map responses between the generated and content images is minimized. 

Formally, let $\mathrm{I}_c$ and $\mathrm{I}_g$ be the content and generated images, respectively. Both $\mathrm{I}_c$ and $\mathrm{I}_g$ share the same image dimensions, and each pixel in $\mathrm{I}_g$ is initialized randomly. Let $F^l_c$ and $F^l_g$ be the feature map responses at VGG-layer $l$ for $\mathrm{I}_c$ and $\mathrm{I}_g$, respectively. Then $\mathrm{I}_g$ is optimized by the minimizing the content loss $\mathcal{L}_{content}(\mathrm{I}_c,\mathrm{I}_g)$ as follows:
\begin{equation}\label{eq:content_loss}
\mathcal{L}_{content} = \frac{1}{2}\sum_l{\left \|F^l_c-F^l_g\right \|_2^2},
\end{equation}
where $\left \| \cdot  \right \|_2^2$ is squared Frobenius norm. Gatys \etal~\cite{gatys2015neural} leveraged the Gram matrix to calculate the correlations between the different filter responses to build a style representation of a given feature map response. The Gram matrix is computed by taking the inner product between the feature maps:
\begin{equation}\label{eq:gram_matrix}
G^l_{ij} = \frac{1}{m^l}\langle F^l_{(i:)},F^l_{(j:)}\rangle,
\end{equation}
where $F^l_{(i:)}$ and $F^l_{(j:)}$ are the $i^\text{th}$ and $j^\text{th}$ vectorized feature maps of VGG-layer $l$, $m^l$ is the number of elements in each map of that layer, and $\langle .,. \rangle$ denotes the inner product. To make the style of the generated $\mathrm{I}_g$ matches the style of a given style image $\mathrm{I}_s$, the difference between the Gram matrices of $\mathrm{I}_g$ and $\mathrm{I}_s$ is minimized as follows:
\begin{equation}\label{eq:style_loss}
\mathcal{L}_{style} = \sum_l{w_l\left \|G^l_s-G^l_g\right \|_2^2},
\end{equation}
where $\mathcal{L}_{style}$ is the style loss, $G^l_s$ and $G^l_g$ are the Gram matrices of $\mathrm{I}_s$ and $\mathrm{I}_g$, respectively, and $w$ refer to scalar weighting parameters to determine the contribution of each layer in $\mathcal{L}_{style}$.

To generate an image $\mathrm{I}_g$ such that the general image content is preserved from $\mathrm{I}_c$ and the texture style statistics are transferred from $\mathrm{I}_s$, the optimization process is jointly performed to minimize the final loss function:
\begin{equation}\label{eq:final_loss}
\mathcal{L} = \alpha \mathcal{L}_{content} + \beta \mathcal{L}_{style},
\end{equation}
where $\alpha$ and $\beta$ are scale factors to control the strength of content reconstruction and style transfer, respectively.

\begin{figure}[t]
\includegraphics[width=\textwidth]{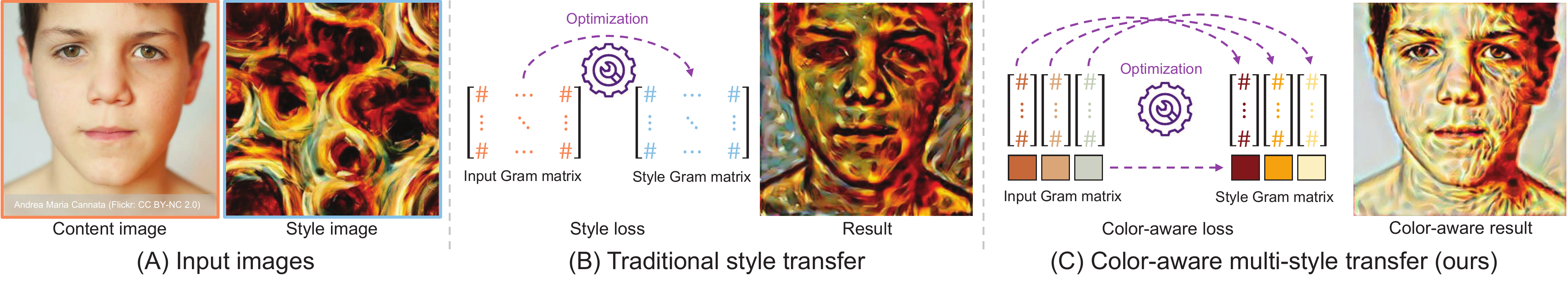}
\vspace{0.5mm}
\caption{Traditional Gram matrix optimization does not consider the correlation between style image's colors and their styles. Therefore, if the style image has more than a single style, like the case shown in (A), this optimization often results in a mixed style as shown in (B). Our method, in contrast, considers the color-style correlation in both images, as shown in (C). Dashed lines in purple refer to the goal of the optimization process.}
\label{fig:motivation}
\end{figure}

From Equation \ref{eq:gram_matrix}, it is clear that Gram matrix measures the correlation of feature channels over \textit{the entire image}. As a result, Gram-based loss in Equation   \ref{eq:style_loss} results in transferring the \textit{averaged global} image style statistics to the generated image. That means if the style image has multiple styles, the traditional Gram-based optimization often fails to sufficiently convey all styles from the style image to the generated image; instead, it would generate images with mixed styles. Figure \ref{fig:motivation} illustrates this limitation. As shown in Figure \ref{fig:motivation}-(A), the style image has more than a single style, which results in unpleasing mixed-style when transferring using traditional Gram-based NST optimization (Figure \ref{fig:motivation}-[B]). Our style transfer result is shown in Figure \ref{fig:motivation}-(C).

Most existing image-optimization NST methods introduce different variations under this main idea of transferring the averaged global image style statistics to the generated one \cite{gatys2015neural, gatys2016preserving, berger2016incorporating, li2017demystifying,  gupta2017characterizing}. However, these methods are restricted with a \textit{single} average style statistics per content and style image pair, and lack artistic controls. While the method of~\cite{risser2017stable} proposed a procedure to control the style of the output image artistically, their procedure requires a tedious human effort that asks the users to annotate semantic segmentation masks and correspondences in both the style and content images.

Unlike other existing methods, we introduce the first Color-Aware Multi-Style (CAMS) transfer method that enables style transfer \emph{locally} based on nearest colors, where multiple styles can be transferred from the style image to the generated one. Our proposed method extracts a color palette from both the content and style images, and \emph{automatically} constructs the region/color associations. The CAMS method performs style transfer, in which the texture of a specific color in the style image is transferred to the region that has the nearest color in the content image. Figure \ref{fig:teaser} shows multiple examples of the generated images (bottom row) from a single input content image with different style images (top row). The regions highlighted in yellow and blue indicate two example styles that were transferred from the style image to regions in the generated image based on the nearest color in the content image.

Our proposed framework allows multi-style transfer to be applied in a meaningful way.  In particular, styles are transferred with associated with colors. By correlating style and colors, we offer another artistic dimension in preserving the content color statistics together with the transferred texture. To further allow artistic controls, we show how our method allows the users to manually select the color associations between the reference style and content image for more transfer options. We believe that our proposed framework and the interactive tool are useful for the research community and enable more aesthetically pleasing outputs. Our source code will be publicly released upon acceptance. 

\section{Our Method} \label{sec:method}

\begin{figure}[t]
\includegraphics[width=\textwidth]{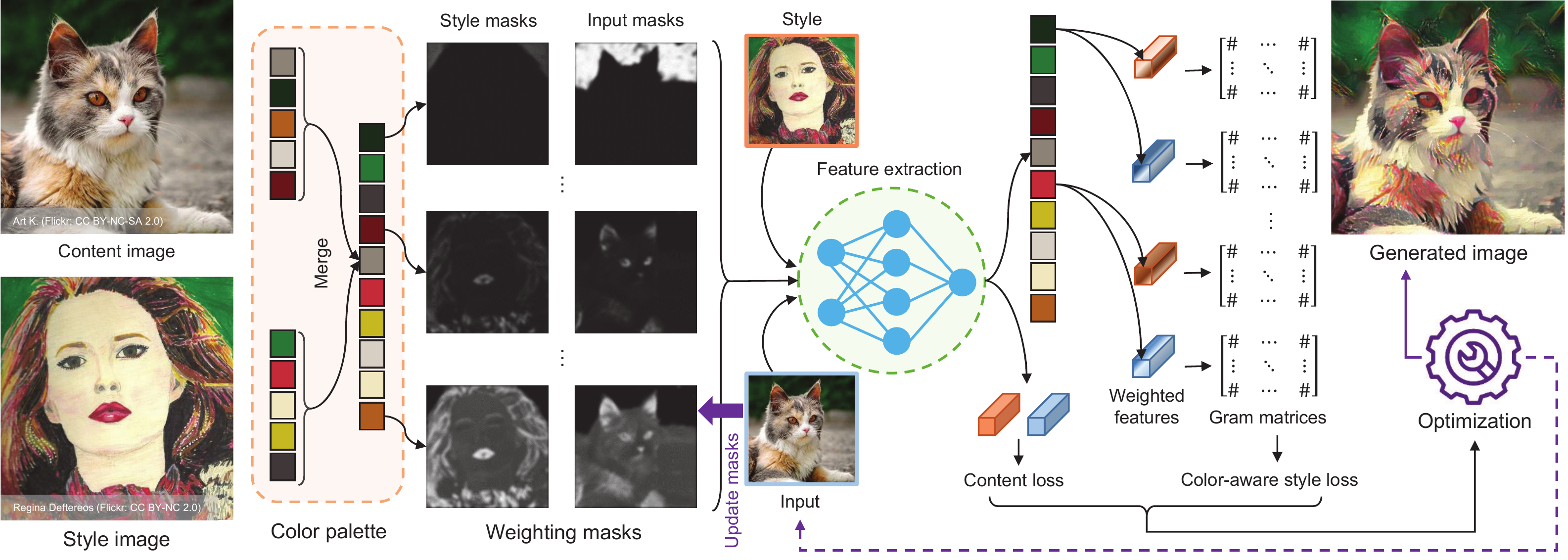}
\vspace{0.5mm}
\caption{Our method proposes to extract color palette from both the content and style images. This color palette is then used to generate color weighting masks. These masks are used to weight the extracted deep feature of both style and input images. This color-aware separation results in multiple Gram matrices, which are then used to compute our color-aware style loss. This loss along with the content loss are used for optimization.}
\label{fig:method}
\end{figure}


Figure \ref{fig:method} illustrates an overview of our method. As shown in the figure, we use color palettes to guide the optimization process. Given two color palettes extracted from the content and style images, respectively. We merge them to generate a single input color palette, $\mathrm{C}$, which is then used to generate a set of color masks. We use these masks to weight deep features of both input and style images from different layers of a pre-trained CNN. This color-aware separation of deep features results in multiple Gram matrices used to compute our style loss during image optimization. In this section, we will elaborate on each step of our algorithm.

\subsection{Mask Generation}
Given an image, $\mathrm{I} \in \mathbb{R}^{n\times m\times3}$, and a target palette, $\mathrm{C}$, our goal is to compute a color mask, $\mathrm{M} \in \mathbb{R}^{n\!\times\!m}$, for each color $\mathrm{t}$ in our color palette $\mathrm{C}$, such that the final mask reflects the similarity of each pixel in $\mathrm{I}$ to color $\mathrm{t}$. We generate $\mathrm{M}$ by computing a radial basis function (RBF) between each pixel in $\mathrm{I}$ and our target color $\mathrm{t}$ as follows:
\begin{equation}\label{RBF}
\mathrm{M}_j = \exp{\left(-\left \| \mathrm{I}_j - \mathrm{t}  \right \|_2/\sigma\right)^2},
\end{equation}
where $\sigma$ is the RBF fall-off factor, $\mathrm{I}_j$ is the $j^{\text{th}}$ pixel in $\mathrm{I}$ and $\mathrm{t}$ is the target color.
Next, we blur the generated mask $\mathrm{M}$ by applying a $15\!\times\!15$ Gaussian blur kernel, with a standard deviation of $5$ pixels. This smoothing step is optional but empirically was found to improve the final results in most cases. 

For each of $\mathrm{I}_s$ and $\mathrm{I}_g$, we generate a set of masks, each of which is computed for a color in our color palette, $\mathrm{C}$. Here, $\mathrm{I}_g$ refers to the current value of the image we optimize, not the final generated image. Note that computing our color mask is performed through differentiable operations and, thus, can be easily integrated into our optimization. 

\subsection{Color-Aware Loss} \label{sec:method.our_loss}

\begin{figure}[t]
\includegraphics[width=\textwidth]{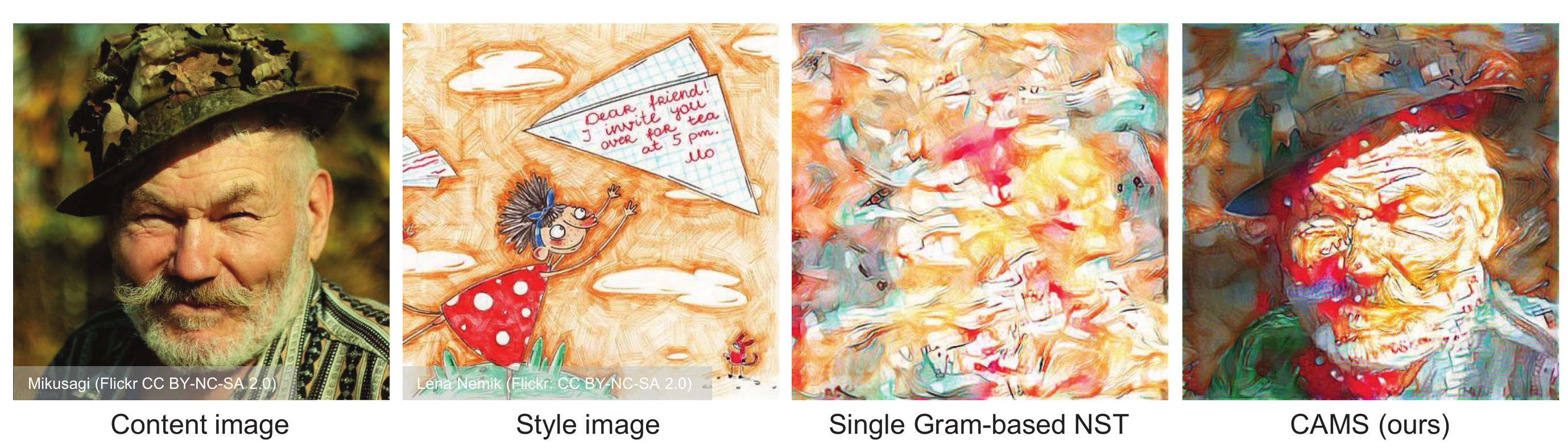}
\vspace{0.5mm}
\caption{In many scenarios, the style image could include multiple styles. Traditional Gram matrix-based optimization (\eg Gatys \etal \cite{gatys2015neural}) cannot capture these styles and as a result it may result in noisy images. In contrast, our color-aware optimization produces more pleasing results while persevering the style-color matching.}
\label{fig:ours_vs_neural_style}
\end{figure}

After computing the color mask sets $\{\mathrm{M}^{(\mathrm{t})}_s\}_{\mathrm{t} \in \mathrm{C}} $ and $\{\mathrm{M}^{(\mathrm{t})}_g\}_{\mathrm{t} \in \mathrm{C}} $ for $\mathrm{I}_s$ and $\mathrm{I}_g$, respectively, we compute two sets of weighted Gram matrices, for $\mathrm{I}_s$ and $\mathrm{I}_g$. According to the given mask weights, each set of weighted Gram matrices captures the correlation between deep features (extracted from some layers of the network) of interesting pixels in the image. This weighted Gram matrix helps our method to focus only on transferring styles between corresponding interesting pixels in the style and our generated image during optimization. 

For a color $\mathrm{t}$ in our color palette $\mathrm{C}$, this weighted Gram matrix, $G^{l(\mathrm{t})}$, is computed as follows:
\begin{equation}\label{weighted_feature}
\hat{F}^l = F^l \odot \mathrm{W}_\mathrm{t},
\end{equation}
\begin{equation}\label{weighted_gram}
G^{l(\mathrm{t})}_{ij} = \frac{1}{m^l}\langle \hat{F}^l_{(i:)},  \hat{F}^l_{(j:)}\rangle,
\end{equation}
\noindent where $\hat{F}^l_{(i:)}$ and $\hat{F}^l_{(j:)}$ are the $i^\text{th}$ and $j^\text{th}$ vectorized feature maps of the network layer $l$ after weighting, $m^l$ is the number of elements in each map of that layer, $\odot$ is the Hadamard product, and $\mathrm{W}_\mathrm{t}$ represents the computed mask for $\mathrm{t}$ after the following processing. First, we linearly interpolate the width and height of our computed mask, for color $\mathrm{t}$, to have the same width and height of the original feature map, $F^l$, before vectorization. Second, we duplicated the computed mask to have the same number of channels in $F^l$.

For each layer $l$ in the pre-trained classification network and based on Equation \ref{weighted_gram}, we compute $\{G^{l(\mathrm{t})}_s\}_{\mathrm{t} \in \mathrm{C}}$ and $\{G^{l(\mathrm{t})}_g\}_{\mathrm{t} \in \mathrm{C}}$ for our style and generated image, respectively. Finally, our color-aware sytle loss is computed as follows:
\begin{equation}\label{eq:color_loss}
\mathcal{L}_{CAMS} = \sum_l^{L_s}\sum_\mathrm{t}{\left \|G^{l(\mathrm{t})}_s - G^{l(\mathrm{t})}_g\right \|_2^2}.\\
\end{equation}


By generating different weighted Gram matrices, our method is able to convey different styles present in the style image, which is not feasible using classic Gram matrix optimization. As shown in Figure \ref{fig:ours_vs_neural_style}, the style image includes different style and textures. NST using Gram matrix (\eg Gatys \etal \cite{gatys2015neural}) fails to capture these multiple styles in the reference style image and produces an unpleasing result as shown in the third column in Figure \ref{fig:ours_vs_neural_style}. In contrast, our color-aware loss considers these styles and effectively transfers them in the generated image as shown in the last column in Figure \ref{fig:ours_vs_neural_style}. For example, the text transferred from the letter (white background) in the style image to the man's white beard in the generated image.

\subsection{Optimization and Implementation Details} \label{sec:method.optimization}

The flow of our method is shown in Algorithm \ref{algorithm}. 
The flow of our method is as follows. First, we initialize each pixel in $\mathrm{I}_g$ with the corresponding one in $\mathrm{I}_c$. Afterward, we generate two color palettes for $\mathrm{I}_g$  and $\mathrm{I}_s$, respectively. We used the algorithm proposed in \cite{chang2015palette} to extract the color palette of each image. The number of colors per palette is a hyperparameter that could be changed to get different results. In our experiments, we extracted color palettes of five colors of each of our content and style images. Then, we merge them to generate the final color palette, $\mathrm{C}$. After merging, the final color palette has at most ten colors, as we exclude redundant colors after merging.

\begin{algorithm}[t]
\scriptsize
\SetAlgoLined
\textbf{Input:} Style image $\mathrm{I}_s \in \mathbb{R}^{n\times m\times3}$, content image  $\mathrm{I}_c \in \mathbb{R}^{n\times m\times3}$, a pre-trained network,  $f$, for image classification, layer indices $L_s$ and $L_c$ for style feature and content features, respectively, and loss term weighting factors, $\alpha$ and $\beta$\\
\KwResult{Generated image $\mathrm{I}_g \in \mathbb{R}^{n\times m\times3}$ that shares styles in $\mathrm{I}_s$ and content in $\mathrm{I}_c$.}
$\mathrm{I}_g = \mathrm{I}_c$\\
$\mathrm{C}_s = \texttt{palette}\left(\mathrm{I}_s\right)$, $\mathrm{C}_g = \texttt{palette}\left(\mathrm{I}_g\right)$\\
$\mathrm{C} = \texttt{merge}\left(\mathrm{C}_s, \mathrm{C}_g\right)$\\
$\{\mathrm{M}^{(\mathrm{t})}_s\}_{\mathrm{t} \in \mathrm{C}} = \texttt{mask}\left(\mathrm{I}_s, \mathrm{C}\right)$, $\{\mathrm{M}^{(\mathrm{t})}_g\}_{\mathrm{t} \in \mathrm{C}} = \texttt{mask}\left(\mathrm{I}_g, \mathrm{C}\right)$\\
$\{F^{S(l)}_s\}_{l \in L_s} = f\left(\mathrm{I}_s, L_s\right), \{F^{C(l)}_s\}_{l \in L_c} = f\left(\mathrm{I}_s, L_c\right)$\\
$\{G^{l(\mathrm{t})}_s\}_{l \in L_s, \mathrm{t} \in \mathrm{C}} = \texttt{weighted\_Gram}\left(\{F^{S(l)}\}_{s_{l \in L_s}}, \{\mathrm{M}^{(\mathrm{t})}_s\}_{\mathrm{t} \in \mathrm{C}} \right)$\\
\While{not converged}{
$\{F^{C(l)}_g\}_{l \in L_c} = f\left(\mathrm{I}_g, L_c\right)$\\
$\mathcal{L}_{content} = \sum_l^{L_c}{\left \|F^{C(l)}_s-F^{C(l)}_g\right \|_2^2}$\\
$\{F^{S(l)}_g\}_{l \in L_s} = f\left(\mathrm{I}_g, L_s\right)$\\
$\{G^{l(\mathrm{t})}_g\}_{l \in L_s, \mathrm{t} \in \mathrm{C}} = \texttt{weighted\_Gram}\left(\{F^{S(l)}_g\}_{l \in L_s}, \{\mathrm{M}^{(\mathrm{t})}_g\}_{\mathrm{t} \in \mathrm{C}}\right)$\\
$\mathcal{L}_{style} = \sum_l^{L_s}\sum_\mathrm{t}{\left \|G^{l(\mathrm{t})}_s - G^{l(\mathrm{t})}_g\right \|_2^2}$\\
$\mathrm{I}_g = \texttt{minimize}\left(\alpha \mathcal{L}_{content} + \beta \mathcal{L}_{CAMS}\right)$\\
$\{\mathrm{M}^{(\mathrm{t})}_g\}_{\mathrm{t} \in \mathrm{C}} = \texttt{mask}\left(\mathrm{I}_g, \mathrm{C}\right)$\\
}
\caption{Color-aware optimization.\label{algorithm}}
\end{algorithm}

After constructing $\mathrm{C}$, we generate color masks $\{\mathrm{M}^{(\mathrm{t})}_s\}_{\mathrm{t} \in \mathrm{C}}$ and $\{\mathrm{M}^{(\mathrm{t})}_g\}_{\mathrm{t} \in \mathrm{C}}$ for $\mathrm{I}_s$ and $\mathrm{I}_g$, respectively. Then, we extract deep features from $\mathrm{I}_s$ and $\mathrm{I}_c$, which represent our target style latent representation and content latent representation, respectively. We adopted VGG-19 net \cite{simonyan2014very} as our backbone to extract such deep features, where we used the $4^{\text{th}}$ and $5^{\text{th}}$ conv layers to extract deep features for the content loss, and the first 5 conv layers to extract deep features for our color-aware style loss. We then construct the weighted Gram matrices, as described in Section \ref{sec:method.our_loss}, using the deep features of style and generated images. The weighted Gram matrices of both generated and style images, and the deep features of generated and content images are used to compute our color-aware style loss (Equation \ref{eq:color_loss}) and the content loss (Equation \ref{eq:content_loss}), respectively. Then, the final loss computed as:

\begin{equation}\label{eq:final_loss_ours}
\mathcal{L} = \alpha \mathcal{L}_{content} + \beta \mathcal{L}_{CAMS},
\end{equation}
\noindent where $\alpha$ and $\beta$ are set to $1.0$ and $10^4$, respectively. After each iteration, we update the color masks of our generated image to track changes in $\mathrm{I}_g$ during optimization. To minimize Equation \ref{eq:final_loss_ours}, we adopted the L-BFGS algorithm \cite{liu1989limited} for 300 iterations with a learning rate of 0.5.

To generate our masks, we have a hyperparameter, $\sigma$, that can be interactively used to control the RBF fall-off, which consequently affects the final result of optimization. Our experiments found that $\sigma = [0.25-0.3]$ works well in most cases.

\begin{figure}[t]
\centering
\includegraphics[width=\textwidth]{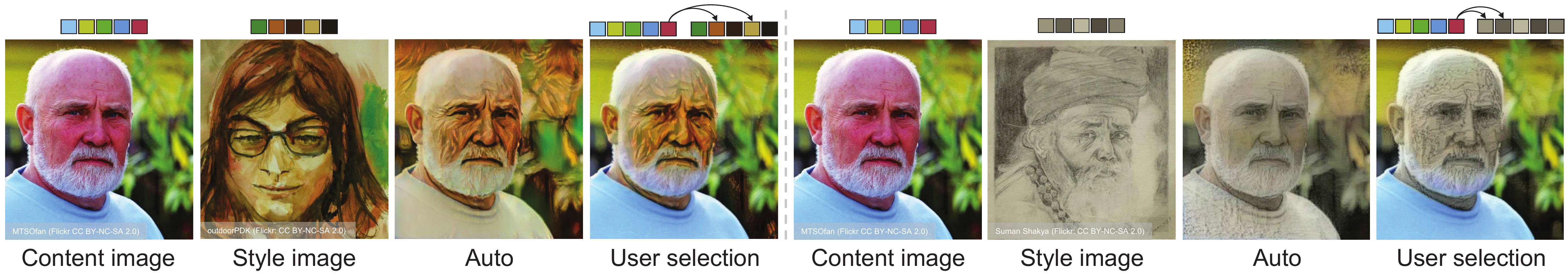}
\vspace{0.1mm}
\caption{Our method allows artistic controls, where the user can manually select color association or discard some colors from the generated palettes. In this figure, we present our results of the auto and user-selection modes for two different reference style images.}
\label{fig:user_selection}
\end{figure}

A nice feature of our method is that it allows more transfer flexibility by enabling the user to determine color associations between our style and content images. To that end, we follow the same procedure explained in Algorithm \ref{algorithm} with the following exception. We do not update color masks of the generated image, $\mathrm{I}_g$, to keep considering the user selection. Figure \ref{fig:user_selection} shows two user cases that reflect the benefit of having our manual user selection tool. As shown in the first case (left), the user associates the reddish color in the content image's color palette to different colors in the style image's color palette. Based on this selection, our generated image has transferred styles based on this color-style correlation. In particular, the change happened only to the style of pixels associated with reddish pixels in the face region. As can also be seen in the second case (right), the transferred style is constrained to those styles associated with selected colors in the style image's color palette.

For the second user case in Figure \ref{fig:user_selection} (bottom row), the auto mode struggled to transfer multiple styles due to the fact that the given style image has a limited range of colors (i.e., only gray-color variations). Such style images, that have limited style options to offer, may result in less appealing outputs. Nevertheless, our manual color-association tool gives the user the flexibility to modify the generated image for more aesthetically pleasing outputs by associating the colors and restricting the modified region as shown in Figure \ref{fig:user_selection} (bottom row).


\begin{figure}[!b]
\includegraphics[width=\textwidth]{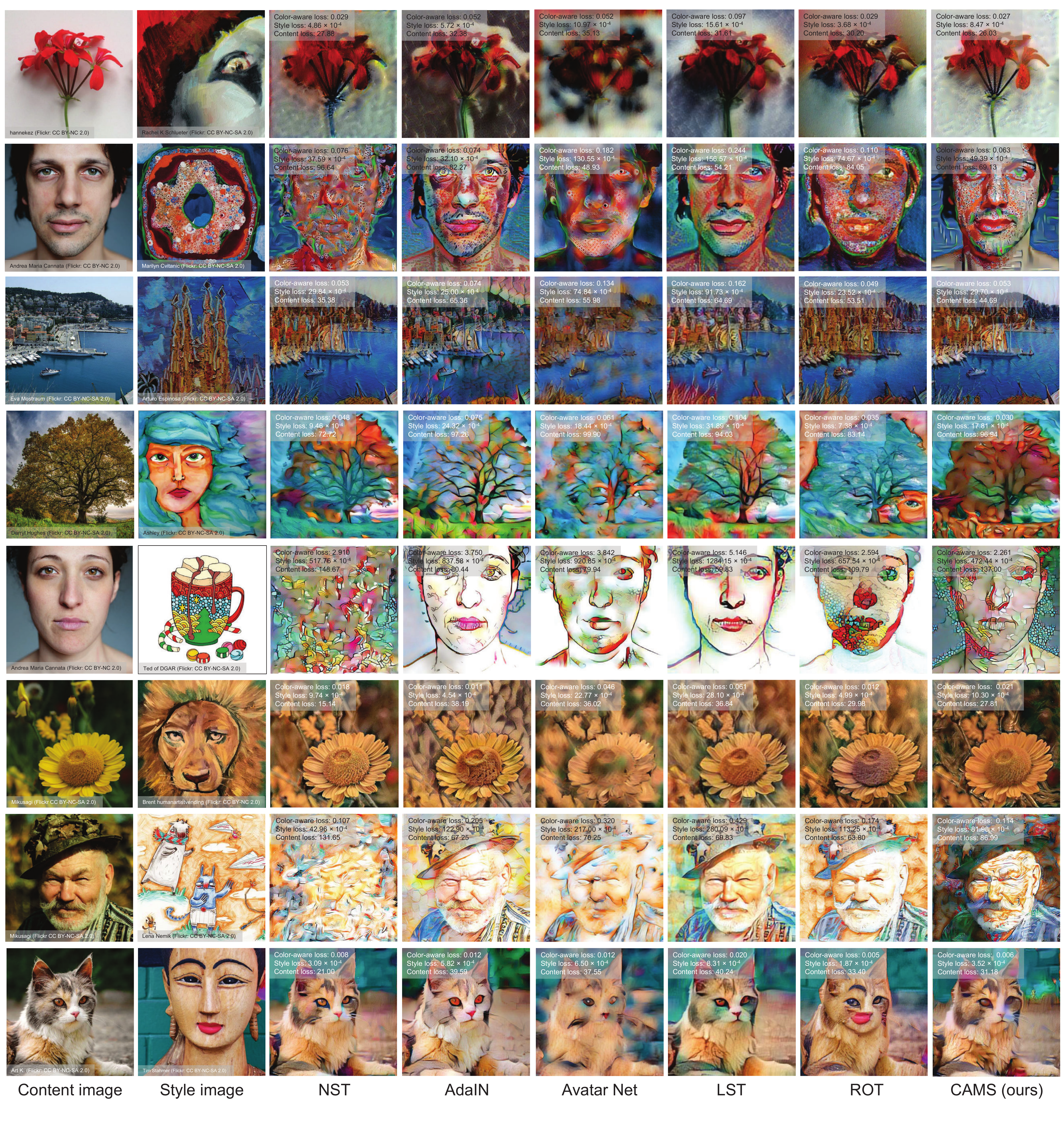}
\vspace{-1mm}
\caption{Qualitative comparisons between our method and other style transfer methods, which are: neural style transfer (NST) \cite{gatys2015neural}, adaptive instance normalization (AdaIN) \cite{huang2017arbitrary}, avatar net \cite{sheng2018avatar}, linear style transfer (LST) \cite{li2018learning}, relaxed optimal transport (ROT) \cite{kolkin2019style}.}
\label{fig:qualitative_fig}
\end{figure}

\section{Evaluation} \label{sec:results}

Evaluating NST techniques is a challenging problem facing the NST community, as indicated in \cite{li2017universal, jing2019neural}. With that said, user studies have been widely adopted in the literature to evaluate subjective results. For the sake of completeness, we conducted an online user study to evaluate our proposed color-aware NST compared with other relevant techniques\footnote{We acknowledge that an in-person user-study within a controlled display environments is preferred, however, due to the covid-19 pandemic we are limited to an online study.}. For each image pair (style/content), we used six different NST methods, which are: neural style transfer (NST) by Gatys \etal~\cite{gatys2015neural}, adaptive instance normalization (AdaIN) \cite{huang2017arbitrary}, avatar net \cite{sheng2018avatar}, linear style transfer (LST) \cite{li2018learning}, and relaxed optimal transport (ROT) \cite{kolkin2019style}. The results of these methods, including ours, were anonymously shown to each subject in a single online webpage, such that the result of each method occupies $\sim$25\% of the screen. For our method, we used the auto-style-transfer mode, where the color matching and optimization were performed automatically as described in Section \ref{sec:method}.

We collected answers from 30 subjects: 18 female and 12 male. The subjects were asked to give a score from one to five for the result of each method anonymously---higher score indicates more appealing result. We evaluated these methods on eight different image pairs. The images were randomly collected from Flick, such that each style image must include more than one style to evaluate our hypothesis of transferring multiple styles from the style image. The image pairs and results -- along with the color-aware loss (Equation \ref{eq:color_loss}), style loss (Equation \ref{eq:style_loss}), content loss (Equation \ref{eq:content_loss}) -- are shown in Figure \ref{fig:qualitative_fig}. Though Figure \ref{fig:qualitative_fig} shows that our method does not always outperform other methods in all loss terms,  our method has a consistent trade-off between the color-aware, style, content losses in most cases, which is confirmed by the user study results (shown in Table \ref{Table:user_study}). As can be seen, 38\% of the total subjects rated the results of our method as high appealing results (score of five). On the other hand, the second best method (i.e., NST~\cite{gatys2015neural}) obtained only 16\% of the votes as five. The study emphasizes the superiority of our proposed method compared with other methods, especially in capturing multiple styles from the style images. Table \ref{Table:quantatitive_flickr} shows the mean value of the color-aware, style, and content losses in our Flickr test set. We also report  the average processing time required by each method on a single NVIDIA GeForce GTX 1080 graphics card in Table \ref{Table:quantatitive_flickr}.

\begin{figure}[!t]

\centering
\includegraphics[width=\textwidth]{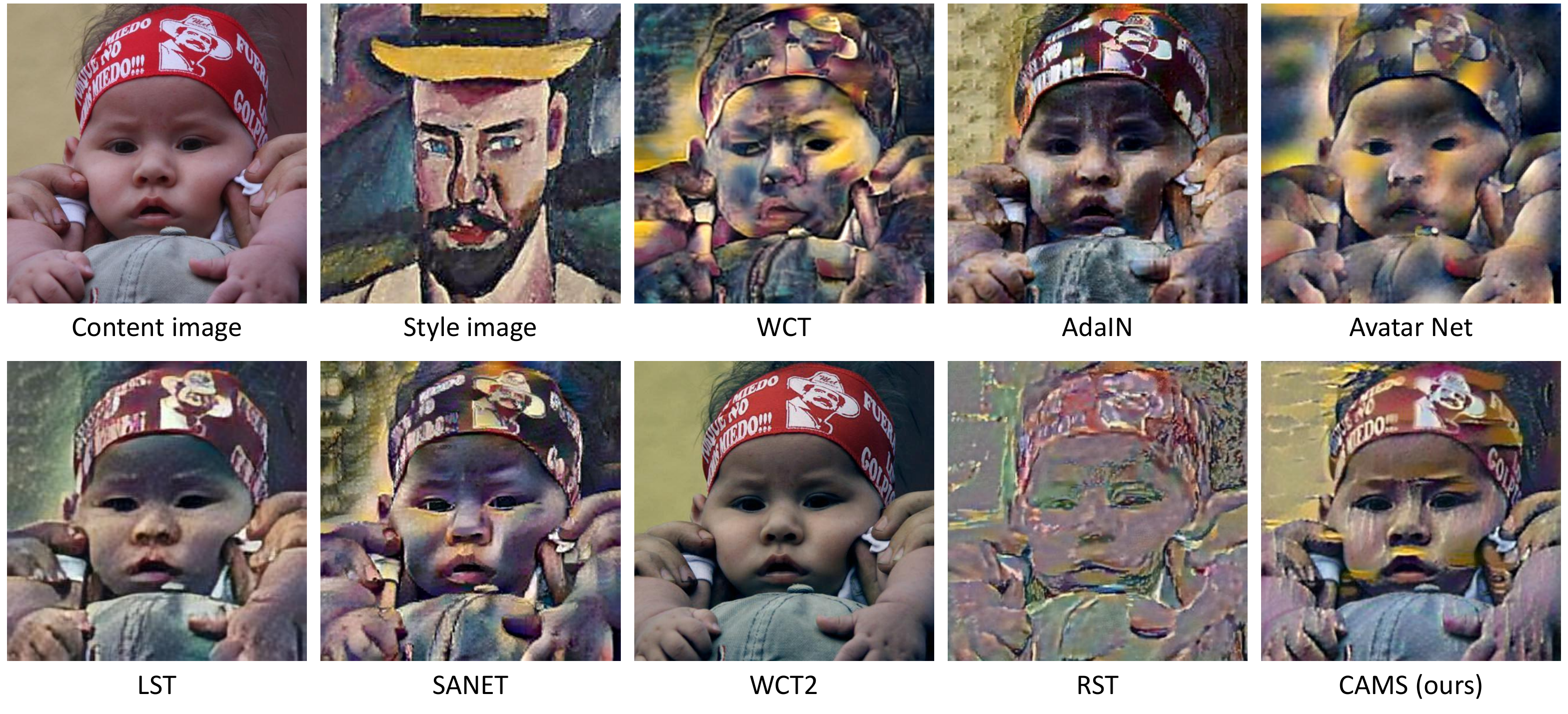}
\vspace{-1mm}
\caption{Qualitative comparisons on the FFHQ $\rightarrow$ WikiArt test set. Shown results of the following methods: whitening and coloring transforms (WCT) \cite{li2017universal}, adaptive instance normalization (AdaIN) \cite{huang2017arbitrary}, avatar net \cite{sheng2018avatar}, linear style transfer (LST) \cite{li2018learning}, arbitrary style Transfer with style-attentional networks (SANET) \cite{park2019arbitrary}, style transfer via wavelet transforms (WCT2) \cite{yoo2019photorealistic} and rethinking style transfer (RST) \cite{kotovenko2021rethinking}.}
\label{fig:face_set}
\end{figure}

In addition to the user-study evaluation, we collected another test set that includes 200 content and style pairs from real portrait face images (randomly selected from the FFHQ dataset \cite{karras2019style}) and painting portrait face images (randomly selected from the face set of the WikiArt dataset \cite{saleh2015large, afifi2021histogan}). We report the color-aware, style, content losses in addition to the FID score \cite{heusel2017gans} achieved by our method and other NST methods \cite{li2017universal, sheng2018avatar,  li2018learning,  yoo2019photorealistic, park2019arbitrary,  kotovenko2021rethinking} in Table \ref{Table:quantatitive_faces}. As shown in Table  \ref{Table:quantatitive_faces}, our method achieves competing results compared to other NST methods across all error metrics; see Figure \ref{fig:face_set} for qualitative comparisons.

\begin{table}[!t]
\begin{minipage}{.5\linewidth}
\caption{Results of user study conducted on 30 subject to evaluate NST methods. Five represents the best aesthetic appealing results.\label{Table:user_study}}
\centering
\resizebox{0.9\textwidth}{!}{
\begin{tabular}{|c|c|c|c|c|c|}
\hline
\multirow{2}{*}{Method}   & \multicolumn{5}{c|}{Rating} \\\cline{2-6} 
& 1 & 2 & 3 & 4 & 5  \\ \hline
NST \cite{gatys2015neural}   & 34\%   & 13\%  & 23\%  & 14\%  & 16\% \\ \hline
AdaIN \cite{huang2017arbitrary}   & 16\%   & 27\%  & 32\%  & 19\%  & 6\%  \\ \hline
Avatar net \cite{sheng2018avatar}  & 50\%   & 27\%  & 17\%  & 4\%   & 2\%  \\ \hline
LST \cite{li2018learning}  & 23\%   & 35\%  & 22\%  & 15\%  & 5\%  \\ \hline
ROT \cite{kolkin2019style}  & 35\%   & 23\%  & 20\%  & 16\%  & 6\%  \\ \hline
CAMS (ours) & 10\%   & 12\%  & 24\%  & 16\%  & \textbf{38\%} \\ \hline
\end{tabular}}
\end{minipage}%
\hspace{1mm}
\begin{minipage}{.45\linewidth}
\centering
\caption{Color-aware, style, and content loss values of NST methods on our collected set from Flickr. The first and second best results are highlighted in yellow (bold) and green, respectively.\label{Table:quantatitive_flickr}}
\resizebox{0.95\textwidth}{!}{
\begin{tabular}{|c|c|c|c|c|}
\hline
Method & Color-aware loss & Style loss & Content loss & Time (secs) \\ \hline
NST \cite{gatys2015neural}  & 0.280 & \cellcolor[HTML]{FFFFC7}\textbf{0.004} &  71.915 & 32.49 \\ \hline
AdaIN \cite{huang2017arbitrary}  & 0.532 & 0.013 & 62.844 & \cellcolor[HTML]{FFFFC7}\textbf{1.05} \\ \hline
Avatar net \cite{sheng2018avatar} & 0.583 & 0.018 & \cellcolor[HTML]{9AFF99}57.963 & 8.82 \\ \hline
LST \cite{li2018learning} & 0.783 & 0.024 & \cellcolor[HTML]{FFFFC7}\textbf{56.413} & \cellcolor[HTML]{9AFF99}3.72 \\ \hline
ROT \cite{kolkin2019style} & 0.376 & 0.011 & 60.986  & 241.11 \\ \hline
CAMS (ours) & \cellcolor[HTML]{FFFFC7}\textbf{0.180} & \cellcolor[HTML]{9AFF99}0.005 & 69.240 & 60.53 \\ \hline
\end{tabular}}
\end{minipage} \hspace{-2mm}
\end{table}

\begin{table}[t]
\caption{Quantitative results (FID \cite{heusel2017gans}, color-aware, style, and content loss values) of NST methods on the FFHQ $\rightarrow$ WikiArt test set. The first and second best results are highlighted in yellow (bold) and green, respectively.\label{Table:quantatitive_faces}}\centering 
\resizebox{0.9\textwidth}{!}{
\begin{tabular}{|c|c|c|c|c|c|c|c|c|}
\hline
Loss/Method &  WCT \cite{li2017universal} & Avatar net \cite{sheng2018avatar} & LST \cite{li2018learning} & SANET \cite{park2019arbitrary} & WCT2 \cite{yoo2019photorealistic} & RST \cite{kotovenko2021rethinking} & CAMS (ours) \\ \hline
Color-aware & 0.009 & 0.019   & 0.020 & \cellcolor[HTML]{9AFF99}0.005 & 0.039 & 0.012 & \cellcolor[HTML]{FFFFC7}\textbf{0.003} \\ \hline
Style $\times 10^{-2}$ & 0.0300 & 0.0790 & 0.0804  & \cellcolor[HTML]{FFFFC7}\textbf{0.0186} & 0.1580 & 0.0570 & \cellcolor[HTML]{9AFF99}0.0218 \\ \hline
Content & 41.839 & 32.445  & 30.871 & 29.085 & \cellcolor[HTML]{FFFFC7}\textbf{2.227} & 44.473 & \cellcolor[HTML]{9AFF99}23.617 \\ \hline
FID & 154.920 &  147.77 & 126.56 & \cellcolor[HTML]{FFFFC7}\textbf{111.58} &  167.54  &  289.60 &  \cellcolor[HTML]{9AFF99}119.93 \\ \hline
\end{tabular}}
\end{table}

\section{Conclusion} \label{sec:conclusion}
We have shown that Gram matrix-based optimization methods often fail to produce pleasing results when the target style image has multiple styles. To fix this limitation, we have presented a color-aware multi-style loss that captures correlations between different styles and colors in both style and generated images. Our method is efficient, simple, and easy to implement, achieving pleasing results while capturing different styles in the given reference image. We also have illustrated how our method could be used in an interactive way by enabling the users to manually control the way of transferring styles from the given style image. Finally, through a user study, we showed that our method achieves the best visually appealing results compared to other alternatives for style transfer. 

\bibliography{ref}
\end{document}